\documentclass[runningheads]{llncs}
\usepackage{placeins}
\usepackage{graphicx}

\usepackage[colorlinks,
            anchorcolor=red,
            citecolor=citecolor, 
            linkcolor=linkcolor,
            ]{hyperref}
            
\usepackage{booktabs}      
\usepackage{amsfonts}       
\usepackage{nicefrac}      
\usepackage{microtype}      
\usepackage{xcolor}                 
\usepackage{url}                         
\usepackage{multirow}
\usepackage{color}
\usepackage{colortbl}
\usepackage{caption}
\usepackage{graphicx}
\usepackage{amsmath}
\usepackage{wrapfig}
\usepackage{lipsum}
\usepackage{algorithm}
\usepackage{algorithmic}
\usepackage{listings}
\usepackage{listing}
\usepackage{xcolor}
\usepackage{framed}
\usepackage{enumitem}
\usepackage{amssymb}
\usepackage{setspace}
\usepackage{epsfig}
\usepackage{subfigure}
\usepackage{mathtools}
\usepackage{verbatim}
\usepackage{booktabs} 
\usepackage{array}
\usepackage{footnote}
\usepackage{tabularx}
\usepackage{booktabs}
\usepackage{todonotes} 
\usepackage{xspace}
\usepackage{colortbl}
\usepackage{bbm}

\definecolor{citecolor}{HTML}{0071BC}
\definecolor{linkcolor}{HTML}{ED1C24}
\def\ie{i.e.}
\def\eg{e.g.}

\begin{document}
\title{Boosting Dermatoscopic Lesion Segmentation via Diffusion Models with Visual and Textual Prompts}

\titlerunning{Diffusion Models with Visual and Textual Prompts}
%


\author{Shiyi Du\inst{1} \and
Xiaosong Wang\inst{1} \and
Yongyi Lu\inst{2} \and
Yuyin Zhou\inst{2,4} \and
Shaoting Zhang\inst{1} \and
Alan Yuille\inst{2} \and
Kang Li\inst{1,3} \and
Zongwei Zhou\inst{2,}\thanks{Correspondence to Zongwei Zhou (\href{mailto:zzhou82@jh.edu}{\textsc{zzhou82@jh.edu}}).}
}

\authorrunning{S. Du et al.}
%

\institute{Shanghai Artificial Intelligence Lab, Shanghai, China
\and
Johns Hopkins University, Baltimore, USA
\and
West China Biomedical Big Data Center, Sichuan University West China Hospital, Chengdu, China
\and
University of California Santa Cruz, Santa Cruz, USA
}

\maketitle           

\begin{abstract}
Image synthesis approaches, \eg, generative adversarial networks, have been popular as a form of data augmentation in medical image analysis tasks. It is primarily beneficial to overcome the shortage of publicly accessible data and associated quality annotations. However, the current techniques often lack control over the detailed contents in generated images, \eg, the type of disease patterns, the location of lesions, and attributes of the diagnosis.  In this work, we adapt the latest advance in the generative model, \ie, the diffusion model, with the added control flow using lesion-specific visual and textual prompts for generating dermatoscopic images. We further demonstrate the advantage of our diffusion model-based framework over the classical generation models in both the image quality and boosting the segmentation performance on skin lesions. It can achieve a 9\% increase in the SSIM image quality measure and an over 5\% increase in Dice coefficients over the prior arts.

\end{abstract}
\section{Introduction}
Image synthesis methods have played an important role in the development of machine vision-based applications as a data augmentation tool to enrich and expand the limited distribution of training data. It is especially helpful for those domains in which sample data and quality annotation are scarce and not cost-effective to obtain, \eg, anonymous driving and medical imaging. Massive research has been conducted in controlling the generated contents for the actual need in model training, from first manipulating the noise parameters $Z$ \cite{DBLP:journals/corr/Goodfellow17}, Conditional GAN (cGAN)~\cite{Isola2016ImagetoImageTW}, to supervised~\cite{Karras2018ASG} and unsupervised~\cite{Rai2018UnpairedIT} image style transfer, to decoupling the style and content parts in images~\cite{Huang2018MultimodalUI}, to latent diffusion model via text2image~\cite{rombach2021highresolution} and most recently diffusion model based ControlNet~\cite{Zhang2023AddingCC}. However, the application of such algorithms in medical image analysis remains limited due to the special characteristics of medical images.  

In most medical diagnosis scenarios, anomaly studies (e.g., image scans with lesions or other abnormalities) are in the minority. Although increasing the number of anomalous samples could potentially help improve the performance of subsequent tasks like segmentation tasks, the generation of such corner cases is not well-controlled and mostly customized at the image level, \eg, the pioneering work in adopting GAN for data augmentation in medical image segmentation~\cite{Shin2018MedicalIS}, adopting cGAN for colon polyp generation~\cite{shin2018abnormal}, diversify the generated image using radiogenomic features~\cite{Xu2020CorrelationVS} and pseudo labels~\cite{lyu2022pseudo}.

Fundamentally, the data sample of such abnormalities remains low in terms of a normal clinical distribution, and it is even hard to obtain enough data for training purposes. Then, the scarce corner cases are often flooded with normal cases or other cases with more common diseases in the set of generated images. There is a critical demand for generating data with desired categories, \ie, specific disease types and disease attributes (shape, location, appearance, severity).  

\begin{figure}[t]
	\centering
        \includegraphics[width=\linewidth]{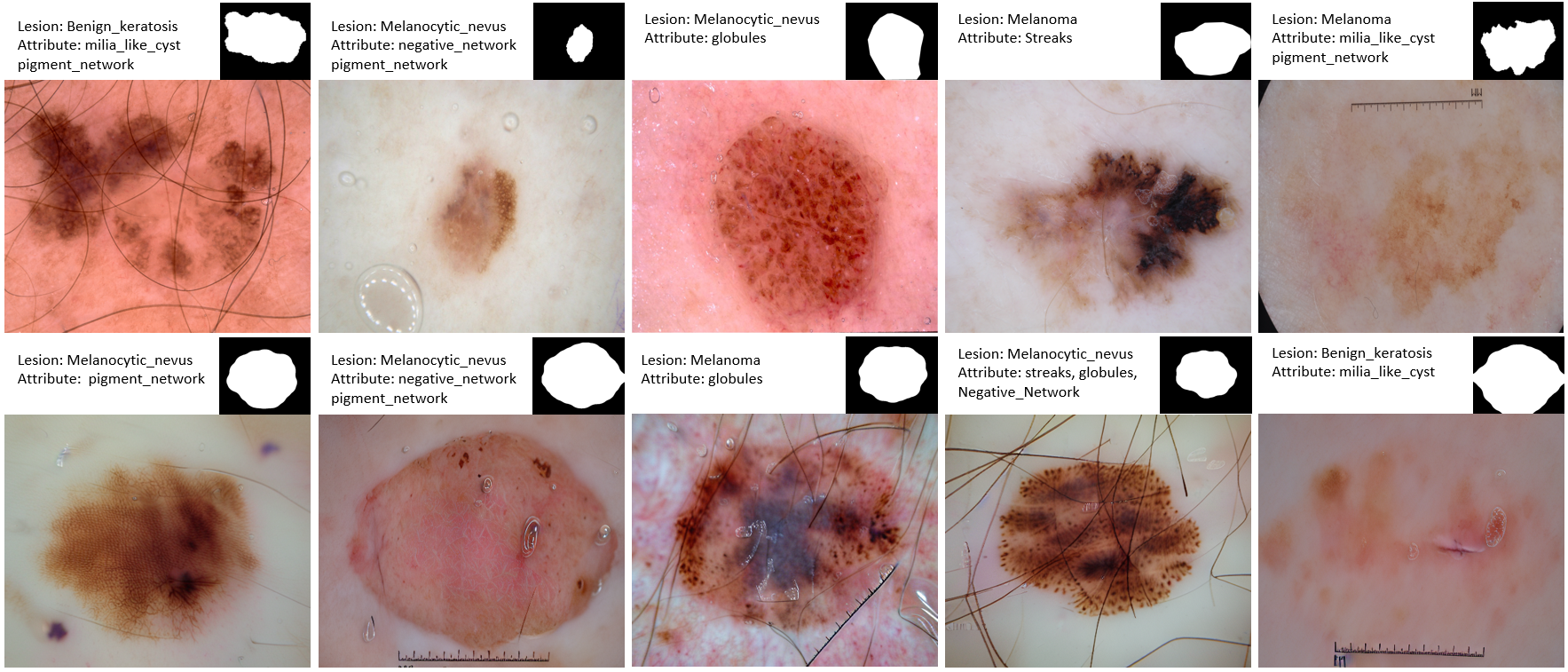}
        \caption{Quiz: which ones are real samples or synthesized by the proposed work? Answers are in the experiments section. Lesion types, attributes, and masks are used as prompts in the training and inference.  }
	\label{fig:sample_real_fake}
\end{figure}

Additionally, the quality and efficiency of the generation of lesion images as well as the transferability of the generation method are essential to improve the performance of subsequent tasks and to apply the method to anomaly images of different organs in different modalities.
The handcrafted methods~\cite{hu2022synthetic,hu2023label,li2023early} are effective when applied to a specific organ and modality, but they are not automated enough and lack universality among various organs and modalities.

In this paper, we propose to use the diffusion models as the backbone to generate skin lesion images, examples are shown in Fig.~\ref{fig:sample_real_fake}. The proposed framework largely leverages the recent work of ControlNet~\cite{Zhang2023AddingCC}, while we attempt to integrate the controllable lesion function (with desired lesion type, attributes in text, and shapes with locations in masks images) into the framework for both the training and inference stage. The correlation could be first learned by linking the visual and textual prompts with the detailed image contents and then prompted during the inference by focusing on rare cases. We also proposed an automatic module to generate lesion shapes and masks. We conducted the experiments and the comparison study (mainly with a classic GAN method, Pix2PixHD~\cite{Wang2017HighResolutionIS}) on a publicly accessible skin lesion dataset, ISIC~\cite{tschandl2018ham10000,DBLP:journals/corr/abs-1902-03368}. The result demonstrates the superiority of the diffusion model-based framework over the classical generation models in
both the image quality and boosting the segmentation performance on
skin lesions. To our best knowledge, we are arguably the first to utilize the diffusion model for skin lesion generation. A PyTorch implementation of our method can be found on our GitHub repository later.

\section{Methods}

\begin{figure}[t]
	\centering
        \includegraphics[width=\linewidth]{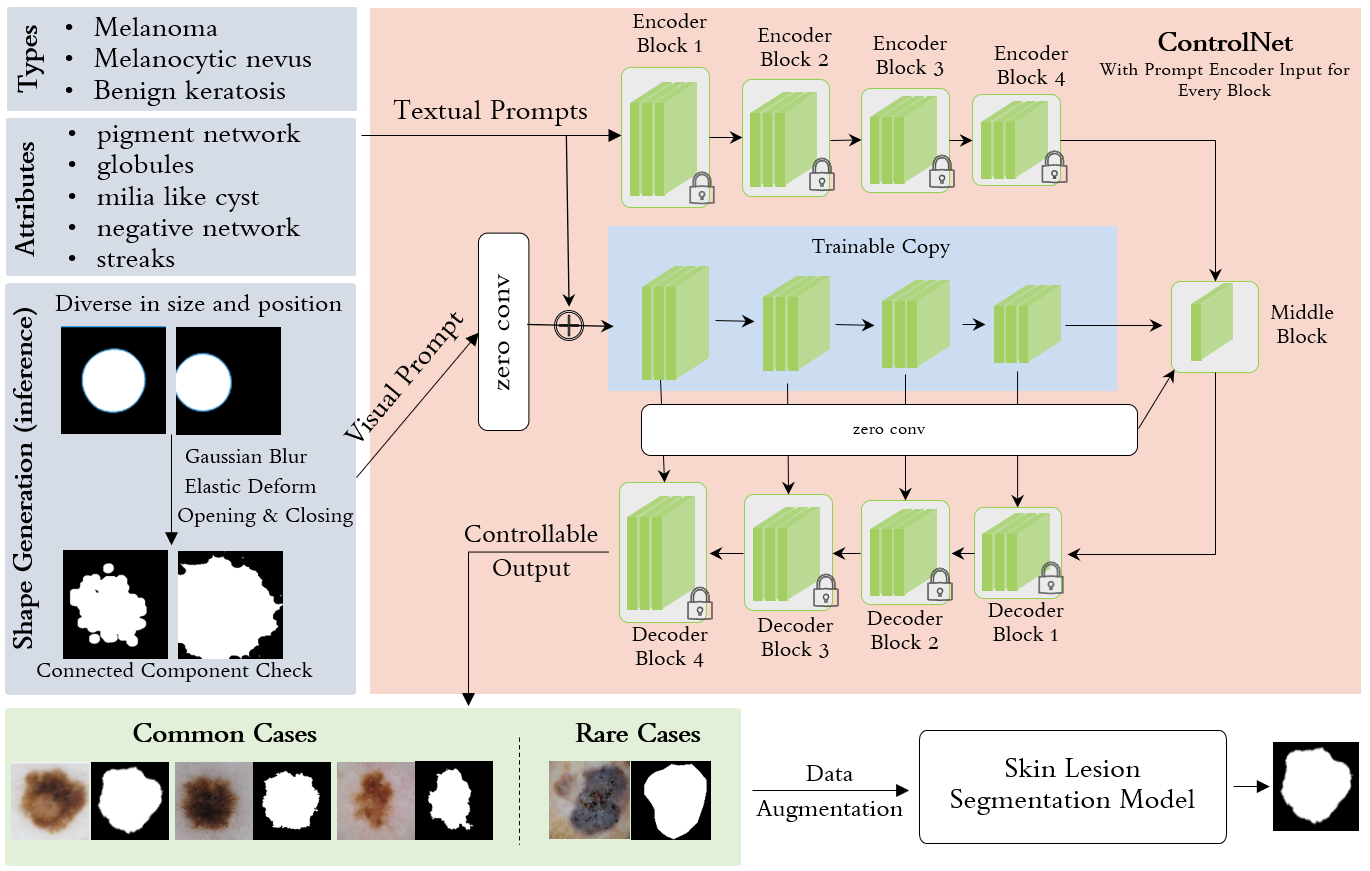}
        \caption{Overview of our visual and textural prompted image generation framework. }
	\label{fig:framework}
\end{figure}

The overall workflow of our method is illustrated in Fig.~\ref{fig:framework}. In the training stage, we utilize the skin images with lesions from the public dataset and their corresponding visual and textual prompts, \ie, the masks of these skin lesions, indicating the shape and location information for the generation, and the lesion types and associated attributes. They are the input to the ControlNet~\cite{Zhang2023AddingCC} for training the lesion image generator, which is imposed on top of the Stable Diffusion model~\cite{rombach2021highresolution}. 
In order to generate the necessary amount of synthetic data to augment the downstream lesion segmentation task, we build an automatic shape generation module, which can produce lesion segmentation masks with a diversity of lesion sizes, shapes, and locations. Theoretically, there is no limit to the number of these generated masks, together with randomly picked textual tags of lesion categories and attributes, that can maximally facilitate the training of downstream lesion segmentation tasks.   

In the following, we first introduce the principle of Denoising Diffusion Implicit Models~\cite{DBLP:journals/corr/abs-2010-02502} as the common diffusion model backbone in \S\ref{ddim}. Then, we explain the diffusion
models with visual and textual prompts in \S\ref{sec_sd_control}. Specifically, we introduce the process of automatically generating lesion shapes in \S\ref{sec_topo_computation}.

\subsection{Backbone Diffusion Model} \label{ddim}
Denoising Diffusion Implicit Model (DDIM) is a recent technique for generative modeling that builds on the common framework of Diffusion Models. DDIM leverages a denoising process to generate high-quality samples from the underlying probability distribution of a dataset.

The key idea behind DDIM is to use a diffusion process to smooth out the noise in an image or data sample, gradually moving it toward the true data distribution. This process is performed by iteratively applying a diffusion equation, where the noise in the data is diffused according to a predetermined schedule of diffusion steps. At each step, the data is corrupted by a certain amount of noise, and the resulting noisy data is used as input for the next step.

The diffusion equation used in DDIM can be written as:

$$\frac{\partial x}{\partial t} = \frac{\beta(t)}{2}\nabla^2 x + \sqrt{\beta(t)}\epsilon,$$

\noindent where $x$ represents the data sample, $t$ is the diffusion time, $\nabla^2$ is the Laplacian operator, $\beta(t)$ is the diffusion coefficient, and $\epsilon$ is Gaussian noise with zero mean and unit variance.

In a DDIM model, we use a denoising objective function that encourages the model to minimize the distance between the diffused noisy data and the original data. Specifically, we use a maximum likelihood approach to learn the parameters of the diffusion process that minimize the negative log-likelihood of the training data. The denoising objective function can be written as:

$$\mathcal{L}(\theta) = \frac{1}{n}\sum_{i=1}^n\sum_{t=0}^T \frac{1}{2\sigma_t^2}|x_{i,t} - \widetilde{x}{i,t}|^2 - \frac{1}{2}\sum{t=0}^T \log \sigma_t^2,$$

\noindent where $\theta$ represents the parameters of the DDIM model, $n$ is the number of data samples, $T$ is the number of diffusion steps, $x_{i,t}$ is the data sample for the $i$-th training example at time $t$, $\widetilde{x}_{i,t}$ is the diffused noisy data, and $\sigma_t$ is the diffusion scale parameter.

One of the advantages of DDIM is that they can generate high-quality images without requiring explicit latent variables or a generator network. Instead, DDIM directly models the data distribution in a continuous manner, making them well-suited for modeling complex distributions with high-dimensional input spaces.

\subsection{Diffusion Models with Multi-modality Prompts} \label{sec_sd_control}

We aim to use diffusion models with multimodal prompts to generate skin images with lesions and then further benefit the downstream segmentation task. First, we train a diffusion model with multimodal prompts for medical images based on Stable Diffusion Models~\cite{rombach2021highresolution} with the multimodal condition, also known as ControlNet~\cite{Zhang2023AddingCC}. 
ControlNet is designed to control the diffusion model by adding additional conditions to facilitate what we call medical image generation through Stable Diffusion Models with multi-modality prompts. As shown in Fig.~\ref{fig:framework}, the network structure is divided into trainable and locked sections in ControlNet. The trainable part is the controllable part that is initialized with the same encoder of the stable diffusion model and then connects the prompts and the detailed generation output. The locked part retains the original parameters of the trainable-diffusion model, so using a small amount of data to bootstrap. Therefore, we can ensure the adapted model learns the desired controlling constraints while retaining the generation capability of the original diffusion model itself. Specifically, the parameters in the upper and lower Encoder and Decoder blocks are locked, and the parameters in the middle blocks are the ``trainable'' ones.

Then, we use the trained model to generate a large number of skin lesion images. At this time, we incorporate the automatically generated skin lesion mask module we mentioned earlier and use the lesion masks generated by this module as input to the trained model to generate the complete skin image with a lesion. According to our subsequent experiments, the segmentation of the model is better to a certain extent as we add more synthetic images to the training.

\subsection{Automatic Lesion Mask Generation} \label{sec_topo_computation}

We experiment with two ways to generate the shape of skin lesions, one of which is the direct transformation according to the existing segmentation masks, and the other is automatically constructing the segmentation masks. However, according to our experimental results, we found that the latter is more flexible and productive in terms of shape diversity and subsequent contribution to the model effect in the segmentation process. Especially, the automated process offers an almost unlimited number of data samples. 
Therefore, we will mainly focus on and discuss more of the automatic lesion shape generation below.

Overall, we generate synthetic images of circles of different sizes and positions and applied some post-processing methods to make the images more realistic. First, a blank canvas is created, then a random point is chosen at the center of the canvas, and a random radius is chosen between a defined threshold and a minimum distance from that point to the edge of the canvas. Next, Gaussian blurring is applied to the image with the inserted circle. Alternatively, the image could be elastically deformed using the elastic deformation library. Finally, morphological on and off operations are performed with elliptical structure elements. The resulting image is labeled using the classic region-growing function to figure out if it has only one connected component (i.e. circle). On the other side, the transformation-based shape generation constructs the shape directly from the original masks, where we operate resizing, rotating, and elastic transforming of the original segmentation masks.

\begin{table}[t]
\centering
\caption{Comparison of image generation between Pix2PixHD and ours.}
\label{tab:my-table}
\begin{tabular}{|l|c|c|c|c|c|}
\hline
\textbf{Model~~~~} & \textbf{~~~~MSE~~~~} & \textbf{~~~PSNR~~~~} & \textbf{~~~~SSIM~~~} \\ \hline
Pix2pixHD & 0.09 & 58.80 & 0.71 \\ \hline
Ours & 0.06 & 61.64 & 0.80 \\ \hline
\end{tabular}

\end{table}
\begin{figure}[htbp]
	\centering
        \includegraphics[width=0.8\linewidth]{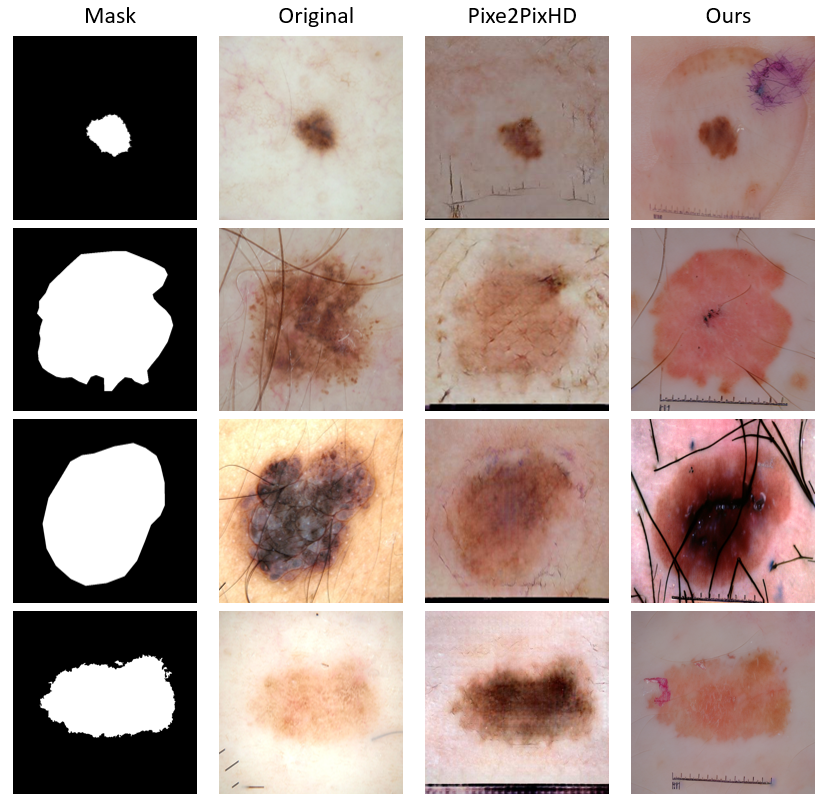}
        \caption{An example of image generation based on masks, from left to right: a skin lesion mask, a Pix2PixHD model generated image, an original image, a diffusion model generated image}
	\label{fig:sample_test_generation}
\end{figure}
\begin{figure}[htbp]
	\centering
        \includegraphics[width=0.6\linewidth]{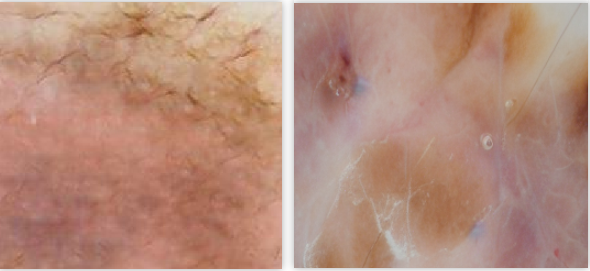}
        \caption{On the left are the texture details generated by the Pix2PixHD model, and on the right are the texture details generated by the Stable Diffusion model with linguistic and visual prompts. }
	\label{fig:sample_texture}
\end{figure}

\section{Experiments}

We conduct the experiments in this study to mainly two aspects of the proposed generation framework, \ie, the quality of generated images and how the controlled generation of sample data (images with associated masks) could benefit the downstream skin lesion segmentation task.

\subsection{Dataset and Experiment Setup} 

The International Skin Imaging Collaboration (ISIC) skin segmentation dataset \cite{tschandl2018ham10000,DBLP:journals/corr/abs-1902-03368} is from the world's largest skin image analysis challenge, hosted by ISIC, a global partnership that has organized the world's largest public repository of dermoscopic images of skin.
Though there are many applications task, e.g., lesion type and attribute classification, the goal of this work is to create a model for segmenting lesion boundaries (same task as ISIC 2018 task 1). There were 2,594 dermoscopic images provided with ground truth segmentation masks. We further cross-correlate the lesion images with the diagnosis and attribute information from ISIC 2017 and 2019 to form the prompts we use for the training of ControlNet. Sample images and associated information are shown in the top row of Fig.~\ref{fig:sample_real_fake} and images on the bottom row of Fig.~\ref{fig:sample_real_fake} are generated via our framework.

We divided the dataset into two parts. Roughly half of the data (1,594 data samples in total) is reserved for the training of the generation model, defined as the G set. The rest (1,000 samples) is employed for the experiments of the lesion segmentation task, defined as the S set. We split the 1,000 images in S set randomly into training, validation, and testing with a ratio of 7:1:2 for evaluating the segmentation performance. We compute Sørensen–Dice coefficient (DSC) for all the segmentation results.

\subsection{Implementation Details}
The diffusion models with visual and text prompts are implemented using Pytorch Lightning based on the codebase provided by~\cite{Zhang2023AddingCC}. The Segmentation and Pix2PixHD model training experiments are carried out on NVIDIA A100 Tensor Core GPUs. To train all models, we use a common segmentation loss function, \ie, Dice loss, since we make it clean and target evaluating the performance gain from additional data samples generated from our framework. The Adam optimizer with a learning rate of 1e-3 is set with a batch size of 256 and a total epoch number of 350 during the training.

\subsection{Results on Image Generation} 
Here, we utilize the testing images of the S set to evaluate the image generation quality. After both the image generation methods, \eg, ours and Pix2PixHD, we use the mask of the original images in the testing of S for the image generation. No prompt is entered, \ie, remaining empty in the prompts for our framework for a fair comparison. In such a way, we can compare the generated image with the original image to see how the methods work in a quantitative manner.

As shown in Table~\ref{tab:my-table}, ours outperforms the Pix2PixHD method by a significant margin. In general, the lesion content and the detailed texture of our generated image are better controlled and produced in comparison to the GAN counterpart. We also illustrate four examples of generation results for both methods in Fig.~\ref{fig:sample_test_generation}.
and enlarged texture regions from both methods are shown in Fig.~\ref{fig:sample_texture}. 
The diffusion model-based image generation can achieve much better generation results with all the details preserved, a clear difference in the degree of texture detail generation as shown in Fig.~\ref{fig:sample_texture}.  

\subsection{Lesion Segmentation Results} 
 As shown in Fig.~\ref{tab:seg-results}, the diffusion model outperforms the Pix2PixHD by a large margin (over 5\%). We also experiment with how different amounts (\eg, 1K, 3K, and 5K) of synthetic data will affect performance. Indeed, more data can benefit the segmentation training, a potential to increase the data amount.
\begin{table}[t]
\centering
\caption{Performance of model lesion segmentation with different generative models adding different amounts of generative images marked by DSC. }
\begin{tabular}{|l|c|c|c||l|c|}
\hline
S + generated images~~                    & \multicolumn{1}{c|}{~S+1K~~~} & \multicolumn{1}{c|}{~S+3K~~~} & \multicolumn{1}{c||}{~S+5K~~~} & SOTA~~~~~~~~~~~ & \multicolumn{1}{c|}{~~S only~~} \\ \hline
Pix2pixHD  & 0.871                     & 0.903                     & 0.912                     & U-Net           & 0.861                  \\ \hline
Ours & \textbf{0.912}            & \textbf{0.913}            & \textbf{0.914}            & DCSAU-Net       & 0.903                  \\ \hline
\end{tabular}
\label{tab:seg-results}
\end{table}

\section{Conclusion}
In this work, we present a diffusion model-based image generation framework with detailed lesion characteristics as the prompts for the lesion image generation task. We demonstrate both quantitatively and qualitatively that the quality of the resulting images is significantly better than the counterpart with the GAN framework, \ie, the popular Pix2PixHD approach. The proposed framework opens the gate of precise data sample generation for multi-tasks, \eg, the segmentation task in this work and possible lesion diagnosis and attributes for image classification. \\
\\
\noindent\textbf{Acknowledgments.}
This work was supported by the Patrick J. McGovern Foundation Award.

\pagebreak

%
%
%
\bibliographystyle{splncs04}
\bibliography{ref,zzhou}

\begin{thebibliography}{10}
\providecommand{\url}[1]{\texttt{#1}}
\providecommand{\urlprefix}{URL }
\providecommand{\doi}[1]{https://doi.org/#1}

\bibitem{DBLP:journals/corr/abs-1902-03368}
Codella, N.C.F., Rotemberg, V., Tschandl, P., Celebi, M.E., Dusza, S.W.,
  Gutman, D.A., Helba, B., Kalloo, A., Liopyris, K., Marchetti, M.A., Kittler,
  H., Halpern, A.: Skin lesion analysis toward melanoma detection 2018: {A}
  challenge hosted by the international skin imaging collaboration {(ISIC)}.
  CoRR  \textbf{abs/1902.03368} (2019), \url{http://arxiv.org/abs/1902.03368}

\bibitem{DBLP:journals/corr/Goodfellow17}
Goodfellow, I.J.: {NIPS} 2016 tutorial: Generative adversarial networks. CoRR
  \textbf{abs/1701.00160} (2017), \url{http://arxiv.org/abs/1701.00160}

\bibitem{hu2023label}
Hu, Q., Chen, Y., Xiao, J., Sun, S., Chen, J., Yuille, A.L., Zhou, Z.:
  Label-free liver tumor segmentation. In: Proceedings of the IEEE/CVF
  Conference on Computer Vision and Pattern Recognition. pp. 7422--7432 (2023)

\bibitem{hu2022synthetic}
Hu, Q., Xiao, J., Chen, Y., Sun, S., Chen, J.N., Yuille, A., Zhou, Z.:
  Synthetic tumors make ai segment tumors better. NeurIPS Workshop on Medical
  Imaging meets NeurIPS  (2022)

\bibitem{Huang2018MultimodalUI}
Huang, X., Liu, M.Y., Belongie, S.J., Kautz, J.: Multimodal unsupervised
  image-to-image translation. In: European Conference on Computer Vision (2018)

\bibitem{Isola2016ImagetoImageTW}
Isola, P., Zhu, J.Y., Zhou, T., Efros, A.A.: Image-to-image translation with
  conditional adversarial networks. 2017 IEEE Conference on Computer Vision and
  Pattern Recognition (CVPR) pp. 5967--5976 (2016)

\bibitem{Karras2018ASG}
Karras, T., Laine, S., Aila, T.: A style-based generator architecture for
  generative adversarial networks. 2019 IEEE/CVF Conference on Computer Vision
  and Pattern Recognition (CVPR) pp. 4396--4405 (2018)

\bibitem{li2023early}
Li, B., Chou, Y.C., Sun, S., Qiao, H., Yuille, A., Zhou, Z.: Early detection
  and localization of pancreatic cancer by label-free tumor synthesis. MICCAI
  Workshop on Big Task Small Data, 1001-AI  (2023)

\bibitem{lyu2022pseudo}
Lyu, F., Ye, M., Carlsen, J.F., Erleben, K., Darkner, S., Yuen, P.C.:
  Pseudo-label guided image synthesis for semi-supervised covid-19 pneumonia
  infection segmentation. IEEE Transactions on Medical Imaging  (2022)

\bibitem{Rai2018UnpairedIT}
Rai, H., Shukla, N.: Unpaired image-to-image translation using cycle-consistent
  adversarial networks (2018)

\bibitem{rombach2021highresolution}
Rombach, R., Blattmann, A., Lorenz, D., Esser, P., Ommer, B.: High-resolution
  image synthesis with latent diffusion models (2021)

\bibitem{Shin2018MedicalIS}
Shin, H.C., Tenenholtz, N.A., Rogers, J.K., Schwarz, C.G., Senjem, M.L.,
  Gunter, J.L., Andriole, K.P., Michalski, M.H.: Medical image synthesis for
  data augmentation and anonymization using generative adversarial networks.
  In: SASHIMI@MICCAI (2018)

\bibitem{shin2018abnormal}
Shin, Y., Qadir, H.A., Balasingham, I.: Abnormal colon polyp image synthesis
  using conditional adversarial networks for improved detection performance.
  IEEE Access  \textbf{6},  56007--56017 (2018)

\bibitem{DBLP:journals/corr/abs-2010-02502}
Song, J., Meng, C., Ermon, S.: Denoising diffusion implicit models. CoRR
  \textbf{abs/2010.02502} (2020), \url{https://arxiv.org/abs/2010.02502}

\bibitem{tschandl2018ham10000}
Tschandl, P., Rosendahl, C., Kittler, H.: The ham10000 dataset, a large
  collection of multi-source dermatoscopic images of common pigmented skin
  lesions. Scientific data  \textbf{5}(1), ~1--9 (2018)

\bibitem{Wang2017HighResolutionIS}
Wang, T.C., Liu, M.Y., Zhu, J.Y., Tao, A., Kautz, J., Catanzaro, B.:
  High-resolution image synthesis and semantic manipulation with conditional
  gans. 2018 IEEE/CVF Conference on Computer Vision and Pattern Recognition pp.
  8798--8807 (2017)

\bibitem{Xu2020CorrelationVS}
Xu, Z., Wang, X., Shin, H.C., Yang, D., Roth, H.R., Milletar{\`i}, F., Zhang,
  L., Xu, D.: Correlation via synthesis: End-to-end image generation and
  radiogenomic learning based on generative adversarial network. In:
  International Conference on Medical Imaging with Deep Learning (2020)

\bibitem{Zhang2023AddingCC}
Zhang, L., Agrawala, M.: Adding conditional control to text-to-image diffusion
  models. ArXiv  \textbf{abs/2302.05543} (2023)

\end{thebibliography}

\end{document}